\documentclass[letterpaper]{article} 
\usepackage{aaai24}  
\usepackage{times}  
\usepackage{helvet}  
\usepackage{courier}  
\usepackage[hyphens]{url}  
\usepackage{graphicx} 
\usepackage{natbib}  
\usepackage{caption} 
\usepackage{algorithm}
\usepackage{algorithmic}
\usepackage{subfigure}
\usepackage{amsmath}
\usepackage{amssymb}
\usepackage{booktabs}
\usepackage{amsfonts}       
\usepackage{nicefrac}       
\usepackage{microtype}      
\usepackage{xcolor}         
\usepackage{enumitem}
\usepackage{makecell}
\usepackage{amsthm}
\usepackage[switch]{lineno}
\usepackage{xcolor} 
\usepackage{multirow}

\usepackage{colortbl}
\definecolor{gray}{RGB}{222,222,222}
\usepackage{newfloat}
\usepackage{listings}
\DeclareCaptionStyle{ruled}{labelfont=normalfont,labelsep=colon,strut=off} 
\floatstyle{ruled}
\newfloat{listing}{tb}{lst}{}
\floatname{listing}{Listing}
\title{Federated Adaptive Prompt Tuning for Multi-Domain Collaborative Learning}
\author{
    Shangchao Su,
    Mingzhao Yang,
    Bin Li\thanks{Corresponding author},
    Xiangyang Xue\footnotemark[1]
}
\affiliations{
   Shanghai Key Laboratory of Intelligent Information Processing\\
School of Computer Science, Fudan University\\
    \{scsu20, mzyang20, libin, xyxue\}@fudan.edu.cn
}

\usepackage{bibentry}

\begin{document}
\maketitle

\begin{abstract}
Federated learning (FL) enables multiple clients to collaboratively train a global model without disclosing their data. Previous researches often require training the complete model parameters. However, the emergence of powerful pre-trained models makes it possible to achieve higher performance with fewer learnable parameters in FL. In this paper, we propose a federated adaptive prompt tuning algorithm, FedAPT, for multi-domain collaborative image classification with powerful foundation models, like CLIP. Compared with direct federated prompt tuning, our core idea is to adaptively unlock specific domain knowledge for each test sample in order to provide them with personalized prompts. To implement this idea, we design an adaptive prompt tuning module, which consists of a meta prompt, an adaptive network, and some keys. The server randomly generates a set of keys and assigns a unique key to each client. Then all clients cooperatively train the global adaptive network and meta prompt with the local datasets and the frozen keys. Ultimately, the global aggregation model can assign a personalized prompt to CLIP based on the domain features of each test sample. We perform extensive experiments on two multi-domain image classification datasets across two different settings --- supervised and unsupervised. The results show that FedAPT can achieve better performance with less than 10\% of the number of parameters of the fully trained model, and the global model can perform well in diverse client domains simultaneously. The source code is available at \url{https://github.com/leondada/FedAPT}.
\end{abstract}
\section{Introduction}
\label{sec:intro}

As privacy protection gains increasing attention, federated learning (FL)~\cite{mcmahan2017communication}, a special machine learning paradigm, becomes more popular. FL has been applied to mobile phone album classification, automatic driving, and medical image analysis. FL enables multiple parties to cooperatively train a global model without sharing their training data. In each round of communication, the server sends the global model to the clients, and each client updates the global model with the private dataset to obtain the local model. The server collects the local models for aggregation, and obtains the global model for the next communication round. In computer vision, FL has been applied to classification~\cite{hsu2020federated,li2021model,yang2023exploring}, detection~\cite{liu2020fedvision,su2023cross}, ReID~\cite{zhuang2021joint,zhuang2020performance}, etc.

\begin{figure}[tb]
\centering
\includegraphics[width=1\linewidth]{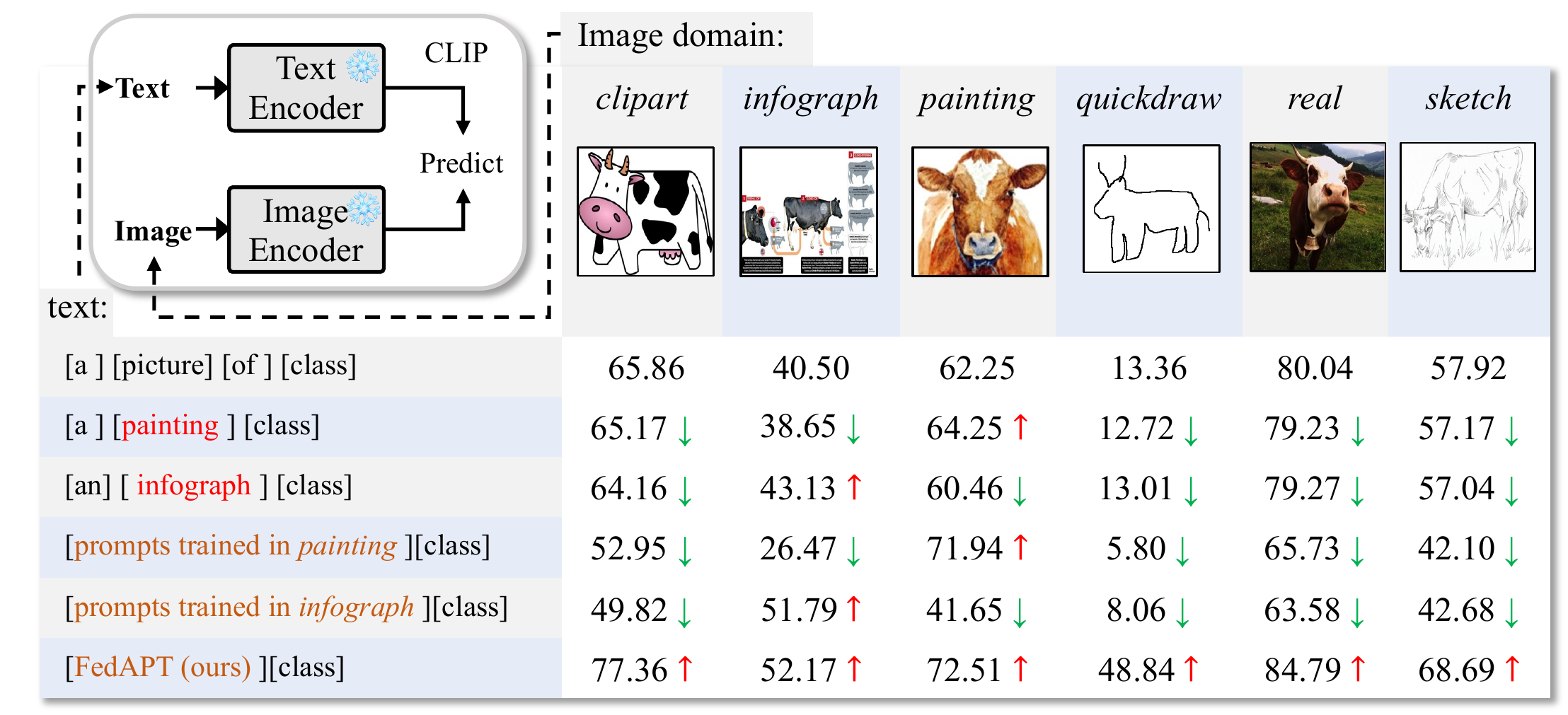}
\caption{The impact of different prompts on performance in multi-domain scenarios. The first three prompts are manually set. One can see that when domain information is included in the prompt (in red), the accuracy of the corresponding domain is improved (red arrow), but that of other domains is decreased (green arrow). The bottom three rows are the learned prompts. On one hand, the prompts learned only in one domain perform significantly worse in other domains. On the other hand, we can obtain prompts through FedAPT that are applicable across all domains.}
\label{intro}
\end{figure}

Currently used FL techniques~\cite{mcmahan2017communication,conf/mlsys/LiSZSTS20,karimireddy2020scaffold} require all model parameters to be updated from the clients and shared on the server. As a result, collaborative learning incurs substantial client training costs, as well as communication costs for both clients and the server. The non-IID problem of various client data also poses significant difficulties for the federated learning algorithm. The performance of the global aggregation model will be impacted when the client datasets come from diverse data domains. Thus, training a more effective global model with fewer communication costs by adjusting only a few parameters remains an open research question.

The emergence of powerful vision-language pre-trained models, such as CLIP~\cite{radford2021learning}, opens up new avenues for solving this problem. CLIP formalizes object recognition into image-text matching, rather than an image-label mapping problem. Text encoder and image encoder are two of the Transformer~\cite{vaswani2017attention} branches that are included in CLIP. About 400 million web-crawled image-text pairs are used for contrastive learning to establish semantic associations between images and texts. Pre-trained CLIP performs admirably in the zero-shot image classification task and demonstrates high transferability in a number of downstream applications. With the powerful representation ability of CLIP, it becomes possible to fine-tune only a few parameters for each client in the federated learning scenarios.

Inspired by the promising capabilities of CLIP, in this research, we propose a federated prompt tuning algorithm, FedAPT, for multi-domain collaborative image classification in cross-silo federated learning. Consider that there are multiple participants in diverse data domains; the ultimate objective is to develop a global classification model through federated prompt tuning that can perform well in all data domains; see Figure~\ref{intro}. We employ a prompt tuning based strategy to provide learnable prompts for the CLIP. To make the global prompt adapt to all domains, we propose an adaptive method to set customized prompts for various test images in order to address the cross-domain challenge.  

The underlying intuition is that images contain domain-specific information. By incorporating this domain-specific information into prompts, we can better guide the pre-trained CLIP model to activate relevant knowledge to that domain, then the classification ability of the corresponding domain can be improved (see the first three rows in Figure~\ref{intro}). The adaptive prompt tuning module includes a meta prompt, an adaptive network, and some frozen keys. Before federated learning, the server randomly assigns each client a frozen key. In local training, each client trains the adaptive network and the prompt using local data and the frozen key.  Considering that the client may not have labels in the real scene, we also design an unsupervised training method for FedAPT. After local training, all updated parameters and prompts are sent to the server for aggregation.  During inference, we leverage an adaptive network to select a specific key for each test image, so as to generate a specific prompt from the meta prompt. 

We conduct extensive experiments on two multi-domain image classification datasets, Office-Caltech10 and DomainNet, across two different settings: supervised and unsupervised. The results show that FedAPT can build a more powerful global model than fully-trained ResNet50 or ViT with less than 10\% of the number of parameters, and demonstrates its ability to perform well in a variety of scenarios. All experiments show that our scheme has significantly improved performance compared with the competitors. 
In general, our contributions are as follows:
\begin{itemize}[noitemsep, leftmargin=*]

\item We utilize CLIP for the first time in federated cross-domain image classification, across two distinct scenarios --- supervised and unsupervised. Our results demonstrate the significant potential of CLIP in federated learning.

\item We propose a federated adaptive prompt tuning framework, FedAPT. With the leverage of keys, FedAPT can provide a personalized prompt for each test image without adding any learnable parameters.

\item Our experiments conducted across different settings on both Office-Caltech10 and DomainNet demonstrate that FedAPT outperforms both the fully-trained models and the existing federated prompt tuning methods.

\end{itemize}

\section{Related Work}

\subsection{Federated Learning}
The idea of aggregating client data distribution information to the cloud can be traced back to SCM~\cite{li2007support}. Recently, FedAvg~\cite{mcmahan2017communication} extends the aggregation idea to neural networks and proposes the federated averaging algorithm. Despite FedAvg's strong performance under the assumption of IID, as the non-IID degree increases (due to the diverse distributions of client datasets), the aggregation model's performance continues to deteriorate, making the non-IID problem a key area of research. Some efforts attempt to enhance the performance of the global aggregation model by improving the optimization algorithm~\cite{conf/mlsys/LiSZSTS20,karimireddy2020scaffold,reddi2020adaptive,wang2020tackling,su2023one} or designing better aggregation methods~\cite{WangYSPK20,singh2020model}, while others~\cite{DBLP:conf/nips/DinhTN20,fallah2020personalized,hanzely2020lower,DBLP:conf/iclr/LiJZKD21,DBLP:conf/icml/LuoWWST22} turn to assigning different models to each client, i.e., personalized FL, such as using local batch normalization~\cite{DBLP:conf/iclr/LiJZKD21}, decoupling the model into personalized and global parts~\cite{DBLP:conf/icml/LuoWWST22}. The proposed FedAPT belongs to the former, i.e., to enhance the global model. 

Some works~\cite{chen2022pre,nguyenbegin} study the impact of pre-training on FL. Unlike the fine-tuning methods, they need to train the complete model. FedPCL~\cite{tan2022federated} studies a setting where each client has $K$ fixed backbones, they fine-tune the projection head by sharing the prototype features of user data. Unlike FedPCL, we do not need to maintain multiple backbones locally, and we do not upload any prototype features.  All in all, different from the existing works, we pay more attention to how to effectively utilize the knowledge in the powerful pre-trained models.

Recently, FedIns~\cite{feng2023towards} shares a similar spirit with our work, both enabling Instance-adaptive Inference. The difference lies in the implementation: FedIns achieves adaptive inference for ViT through learnable Keys and SSF pool, while our approach achieves adaptive inference for CLIP through randomly initialized keys and a learnable query network coupled with prompts.

\subsection{CLIP and Prompt Tuning}
The contrastive vision-language pre-trained model CLIP~\cite{radford2021learning} transforms image recognition into an image-text matching problem, freeing the object recognition task from human-annotated data, such that a huge amount (400M) of noisy image-text pairs can be used for training. The pre-trained model can be transferred to multiple datasets. Google releases ALIGN~\cite{jia2021scaling} to expand CLIP's training data to 1.8B, which achieves better results. Thanks to its powerful representation capability, CLIP has also been successfully applied to a variety of downstream visual tasks, such as RegionCLIP~\cite{zhong2022regionclip} and DetCLIP~\cite{yao2022detclip} for object detection, MedCLIP~\cite{wang2022medclip} for medical image analysis, CCR-CLIP~\cite{yu2023chinese} for text recognition. 

Prompt~\cite{liu2021pre} adds additional words or sentences to the input text of the pre-trained language model, to make the pre-trained model better handle downstream tasks. The succeeding works~\cite{li2021prefix,liu2021p} treat the prompts as continuous vectors in order to learn new knowledge in the process of fine-tuning, which is called Prompt Tuning. In CLIP's follow-up work, other researchers~\cite{zhou2022learning,zhou2022conditional} tried to improve prompt tuning in CLIP. The work most relevant to ours is CoCoOp~\cite{zhou2022conditional}, which trains prompts conditioned on visual features. However, it is designed for centralized training and heavily relies on high GPU computational power, making it hard to apply to clients in federated learning. In contrast, in this paper, the training of conditions and prompts are decoupled, significantly accelerating the training process.

\subsection{CLIP in Federated Learning}
In the field of federated image classification, work related to CLIP is rare. PromptFL~\cite{guo2022promptfl} extends CoOp~\cite{zhou2022learning} to the federated learning scenarios, averaging multiple prompts on the server. PromptFL shows the potential application of prompt tuning for CLIP in FL, however, it lacks optimizations tailored for multi-domain scenarios. In comparison, FedAPT proposes an adaptive prompt tuning method for multi-domain image classification. FedCLIP~\cite{lu2023fedclip} adds an adapter consisting of two fully connected layers to the end of the image encoder. FedTPG~\cite{qiu2023text} proposes a prompt generator with better generalization performance on unknown categories. 
However, they do not generate personalized prompts for the images based on the domain information.

\section{Preliminaries}
\subsection{Notations}


Consider a federated learning scenario, such as autonomous driving, where multiple cars are deployed in $K$ regions (domains). Data from the same region exhibits similar styles, while data from different regions possesses distinct styles. Each region has $N_{dom}$ clients. We assume that each client has sufficient computing power to train the prompts. In supervised setting, $\mathcal{D}_n=\{\mathbf{x}_i,y_i\}_{i=1}^{m_n}, n=1,\cdots,N$ denote the local datasets, $m_n$ is the number of images in the $n$-th client. In unsupervised setting,  $\mathcal{D}_n=\{\mathbf{x}_i\}_{i=1}^{m_n}, n=1,\cdots,N$. Images in diverse domains have different styles, see Figure~\ref{intro}. Note that there may exist domain and category divergences across different clients at the same time, the divergence of category refers to the different label distribution of different clients. Let $\mathcal{I}(\cdot)$ and $\mathcal{T}(\cdot)$ denote the image encoder and the text encoder. To reduce the calculation and communication costs, we freeze the parameters of the two encoders and perform the training task by fine-tuning the prompts.

\subsection{Prompt Tuning}
Let $\mathbf{t}^c$ be the word embedding of the $c$th-class, the class text could have the form `a picture of a dog'. Then the prediction of CLIP for image classification is:

\begin{linenomath}\begin{align}
p(c \mid \mathbf{x})=\frac{\exp \left(\cos \left(\mathcal{T}(\mathbf{t}^c),\mathcal{I}(\mathbf{x})\right) / \tau\right)}{\sum_{i=1}^C \exp \left(\cos \left(\mathcal{T}(\mathbf{t}^i), \mathcal{I}(\mathbf{x})\right) / \tau\right)}
\end{align}
\end{linenomath}
where $C$ is the number of classes, $\tau$ is a temperature parameter learned by CLIP.

To fine-tune the model, one method is to manually set prompts, such as replacing `a picture of a dog' with `a picture of a cute dog'.  However, in this way, because the change of words will directly affect the performance of the model, we need to find the most appropriate words. Another method is prompt tuning~\cite{li2021prefix,liu2021p,zhou2022learning}. Let  $\mathbf{p}^c=[\boldsymbol{v}^c_1, \boldsymbol{v}^c_2, \ldots, \boldsymbol{v}^c_M]$ be the prompt. The input of the $c$-th class for the text encoder is $(\mathbf{p}^c;\mathbf{t}^c)$, where 
$\{\boldsymbol{v}^c_i\}^{M}_{i=1}$ are learnable prompts for the $c$-th class. $\boldsymbol{v}^c_i$ has the same dimension as word embedding $\mathbf{t}^c$, $M$ is the number of learnable prompts. 


\begin{figure*}[tb]
\centering
\includegraphics[width=0.9\linewidth]{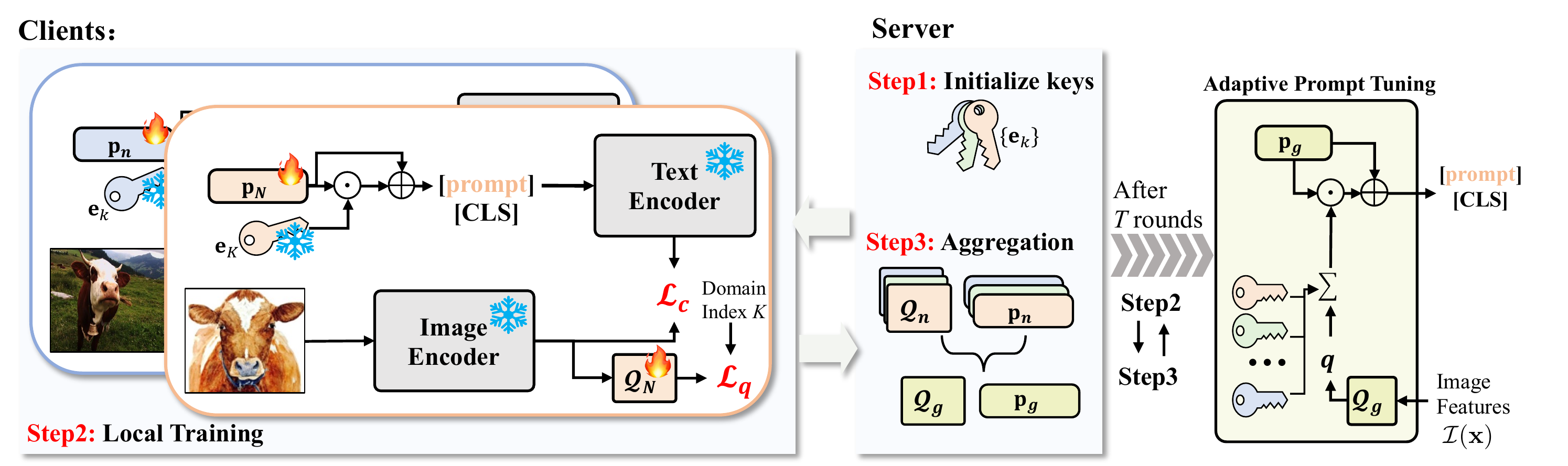}
\caption{Overview of FedAPT. The federated training process is as follows: \textbf{Step1}: The server first randomly generates a set of random keys and assigns each client a frozen key. \textbf{Step2}: In local training, the key is used to give a constraint to the prompt. In addition, each client trains the adaptive network $\mathcal{Q}_n$ with local data. \textbf{Step3}: After local training, the server averages the prompts and adaptive networks learned by the clients and sends the results back to the clients. Repeat Steps 2$\sim$3 for multiple rounds, we can obtain $\mathbf{p}_g$ and $\mathcal{Q}_g$ to establish our target global model which can generate a personalized prompt according to image features. }
\label{framework}
\end{figure*}

\subsection{Federated Learning with Prompt}
\label{fltun}
The concept of FL is introduced in FedAvg~\cite{mcmahan2017communication} to collaboratively train a global model that works for each participant. A straightforward way to apply CLIP's prompt tuning to the FL field is that each client trains its own prompt and uploads the prompt to the server. The server averages the prompts of all clients and distributes them to the clients. 
Let $\mathbf{p}^c_{n}$ denotes the prompt for the $c$-th class in the $n$-th client. For the sake of brevity, we omit $c$ and use $\mathbf{p}_n$ to denote the prompt in the $n$-th client. Let $\mathbf{p}_g$ denotes the prompt of the global model. Then the global objective is:
\begin{linenomath}\begin{align}
\mathbf{p}_g = \arg\min _{\mathbf{p} \in \mathbb{R}^d} \frac{1}{N} \sum_{n=1}^N \mathbb{E}_{\mathbf{x},y\sim \mathcal{D}_n}[\ell(\mathcal{T}(\mathbf{p}) ; \mathcal{I}(\mathbf{x}))]
\end{align}\end{linenomath}
where $\ell$ is the loss of the $n$-th client. Although this approach can handle scenarios where multiple clients are IID, it is still difficult for a single global prompt to process clients from diverse data domains.

\section{Method}

The overview of the proposed FedAPT is in Figure~\ref{framework}. We focus on the problem setting of one-model-fits-all, that is, learning a powerful global model to adapt to the data distribution across all domains. With the parameters of the CLIP frozen, we introduce an additional adaptive prompt tuning (APT) module for the global model. APT accepts image features as input and outputs a specific prompt for each sample. The learnable parts of APT are the prompt and the adaptive network, which are obtained through the federated training of multiple clients.

\subsection{Adaptive Prompt Tuning}
The input of the text encoder in PromptFL always uses a fixed prompt for all samples: $\mathbf{p}_g\in \mathbb{R}^{C\times s\times d}$, which is the learned prompt by federated averaging, $C$ is the number of classes, $s$ is the length of prompt, $d$ is the embedding dimension. However, under the setting in this paper, each client may come from a specific domain, and there are obvious domain divergences among the images. A fixed prompt is no longer suitable for images in all domains. It is necessary to design a new form. An intuitive way is to introduce personalized prompts into the input of the text encoder for each domain:
\begin{linenomath}\begin{align}
\mathbf{p}_g +  \mathbf{p}_{p_k}
\label{ddddd}
\end{align}\end{linenomath}
where $\mathbf{p}_{p_k}$ encodes the personalized domain knowledge of the $k$-th domain. Unfortunately, under the goal of one-model-fits-all, we need a unified global model, rather than adding personalized parameters $\mathbf{p}_{p_k}$ for each client. Then a compromise improvement for Eq.~\ref{ddddd} is to keep all $\mathbf{p}_{p_k}$ in the global model, then filter prompts based on the image features:
\begin{linenomath}\begin{align}
 \mathbf{p}_g + f( [\mathbf{p}_{p_1},\cdots,\mathbf{p}_{p_K}],\mathcal{Q} (\mathcal{I}(\mathbf{x})))
\end{align}\end{linenomath}
where $\mathbf{p}_{p_k}$ is the personalized prompt for the $k$-th domain, $f(\cdot)$ is a filter function, $\mathcal{Q}$ is an adaptive network that is used to provide auxiliary information for filtering. However, this operation needs to retain a number of $\mathbf{p}_{p_k}$, thus introducing additional parameters in the global model.

Finally, we adopt an adaptive prompt tuning approach. Let $\mathcal{P}$ denotes the output of the APT module, the input for this module is the image features $\mathcal{I}(\mathbf{x})$.
We use the key vector $\mathbf{e}_k\in\mathbb{R}^{s\times d}$ to encode personalized domain knowledge into $\mathbf{p}_g$, and decode it when necessary. $\mathcal{P}$ in the global model is:
\begin{linenomath}\begin{align}
\mathcal{P}(\mathcal{I}(\mathbf{x})) = \mathbf{p}_g + \mathbf{p}_g\odot \sum_{k=1}^K q_k \mathbf{e}^{\prime}_k
\label{eq.pg}
\end{align}\end{linenomath}
where $q_k= \mathcal{Q}_g (\mathcal{I}(\mathbf{x}))_k$, $\sum_k q_k =1$, $\odot$ represents the element-wise multiplication, $\mathbf{e}^{\prime}_k\in \mathbb{R}^{C\times s\times d}$ is copied from $\mathbf{e}_k$ for $C$ times to match the dimension of $\mathbf{p}_g$, and $\mathcal{Q}_g$ is the adaptive network with parameters $\phi$, which can give a one-hot or soft-membership vector of $K$ dimensions. In the $n$-th client from the $k$-th domain, the prompt for the text encoder in the training process is:
\begin{linenomath}\begin{align}
\mathbf{p}_n + \mathbf{p}_n\odot \mathbf{e}^{\prime}_k
\label{eq.pl}
\end{align}\end{linenomath}
where $\mathbf{p}_n $ is initialized by $\mathbf{p}_g$, $\mathbf{e}^{\prime}_k$ is frozen. Through the design of Eqs.\ref{eq.pg} and \ref{eq.pl}, we do not need to introduce additional learnable parameters $\mathbf{p}_{p_k}$, but use the unified $\mathbf{p}_{g}$, which is named ``meta prompt".  The objective for the meta prompt is:
\begin{linenomath}\begin{align}
    \nonumber
\mathbf{p}_{g} = \arg\min _{\mathbf{p}} \frac{1}{N} \sum_{n=1}^N \mathbb{E}_{\mathbf{x}\sim\mathcal{D}_n}[\ell(\mathcal{T}(\mathbf{p}+\mathbf{p}\odot \mathbf{e}^{\prime}_n) ; \mathcal{I}(\mathbf{x}))]
\end{align}\end{linenomath}

\subsection{Federated Training}
\label{trainsec}

To enable multiple clients to train APT cooperatively, there are three steps. At the beginning of federated training, the server randomly initializes the keys and sends them to different clients (\textbf{Step 1}). 

\textbf{Step2: Local Training.} To evaluate the capabilities of CLIP and FedAPT in various FL settings, we consider two different client settings:

In the supervised setting, each client updates the global $\mathbf{p}_g$ and $\mathcal{Q}_g$ with the local data respectively to obtain $\mathbf{p}_n$ and $\mathcal{Q}_n$. The prediction of the $n$-th local model is:
\begin{linenomath}\begin{align}
\nonumber
p(c \mid \mathbf{x})=\frac{e^{\cos \left(\mathcal{T}(\mathbf{p}^c_n + \mathbf{p}^c_n\odot \mathbf{e}_k ),\mathcal{I}(\mathbf{x})\right) / \tau}}{\sum_{i=1}^C e^{\cos \left(\mathcal{T}(\mathbf{p}^i_n + \mathbf{p}^i_n\odot \mathbf{e}_k ), \mathcal{I}(\mathbf{x})\right) / \tau}}
\end{align}\end{linenomath}
where $k$ is the domain index of the client. Then the cross entropy loss for image classification is:
\begin{linenomath}\begin{align}
\label{lc}
\mathcal{L}_c= \frac{1}{\left|\mathcal{D}_n\right|} \sum_{\mathbf{x}_i, y_i\sim \mathcal{D}_n} -\log p\left(y_i \mid \mathbf{x}_i, \mathbf{p}_n\right)
\end{align}\end{linenomath}
%

In an unsupervised setting, we design a two-stage training approach. In the first stage, we augment the unlabeled data (same augment strategy as SimCLR~\cite{chen2020simple}) and utilize pseudo-labels with high confidence from the original data as supervised information to perform self-training on the augmented data. The training loss is:
\begin{linenomath}\begin{align}
\label{pseudo}
\mathcal{L}_{pseudo}= \frac{1}{\left|\mathcal{D}_n\right|} \sum_{\mathbf{x}_i \sim \mathcal{D}_n} -\log p\left(\hat{y}_i \mid \hat{\mathbf{x}}_i, \mathbf{p}_n\right)
\end{align}\end{linenomath}
where $\hat{y}_i$ is the pseudo one-hot label of $\mathbf{x}_i$ predicted by the original CLIP, $\hat{\mathbf{x}}_i$ is the augmented version of $\mathbf{x}_i$. After the first stage, we can get a coarse prompt.

In the second stage, use $\mathbf{z}_i$ to represent the prediction score vector of $\mathbf{x}_i$, we hope to compare the outputs among different samples. For inter-sample comparison, motivated by AaDLoss~\cite{yang2022attracting} in domain adaptation, for a certain sample, the logits of $K$-nearest neighbor samples with similar image features should be as similar as possible, otherwise, there should be differences. Then the loss is:
\begin{linenomath}\begin{align}
\label{inter}
\mathcal{L}_{inter}= \frac{1}{\left|\mathcal{D}_n\right|} \sum_{\mathbf{x}_i \sim \mathcal{D}_n} (-\sum_{j \in \mathcal{C}_i} \mathbf{z}_i^\top \mathbf{z}_j+\lambda \sum_{m \in \mathcal{B}_i} \mathbf{z}_i^\top \mathbf{z}_m)
\end{align}\end{linenomath}
where $ \mathcal{C}_i$ contains $K$-nearest neighbors, $ \mathcal{B}_i$ contains the rest samples. A feature bank and a score bank are used to efficiently implement this loss.

As for intra-sample comparison, we need that the prediction of a sample is close to that of its augmented counterpart:
\begin{linenomath}\begin{align}
\label{intra}
\mathcal{L}_{intra}= \frac{1}{\left|\mathcal{D}_n\right|} \sum_{\mathbf{x}_i \sim \mathcal{D}_n} KL(\hat{\mathbf{z}}_i \|\mathbf{z}_i)
\end{align}\end{linenomath}
where $\hat{\mathbf{z}}_i$ is the predict score of the augment data, $\mathbf{z}_i$ is extracted from the score bank. Then the overall loss in the second stage is $\mathcal{L}_{c}=\mathcal{L}_{inter}+\mathcal{L}_{intra}$.

The adaptive network $\mathcal{Q}_n(\cdot)$ is a fully-connection layer with parameters $\phi_n$, which tries to classify domain information from the features of images. Each client in the $k$-th domain trains its own adaptive network with the following cross-entropy loss:
\begin{linenomath}\begin{align}
\label{lq}
\mathcal{L}_q= \frac{1}{\left|\mathcal{D}_n\right|} \sum_{\boldsymbol{x} \in \mathcal{D}_n} -\log p\left(k \mid \mathbf{x}_i, \phi_n\right)
\end{align}\end{linenomath}
The whole loss function for local training is $\mathcal{L}_c+\mathcal{L}_q$.

Note that training $\mathcal{Q}_g(\cdot)$ is a special FL problem, because each client can only access the data of its own domain. In this paper, we find that in the multi-domain scenario, we can obtain good enough $\mathcal{Q}_g(\cdot)$ by directly using cross-entropy for local training and federated averaging for aggregation. To further improve the effect of $\mathcal{Q}_g(\cdot)$, it is necessary to use special training methods, such as FedAws~\cite{yu2020federated}. In addition, directly uploading the data prototype~\cite{tan2022federated,peng2019federated}, can also train the $\mathcal{Q}_g(\cdot)$ in the server, but in order to prevent privacy disclosure, we do not upload any data features or prototypes.

\textbf{Step3: Aggregation}. After the local training step, the $n$-th client uploads $\mathbf{p}_n$ and  $\phi_n$ to the server. Due to $ \phi_n\in\mathbb{R}^{d_{img}\times K}$, $d_{img}$ is the dimension of each image feature, which is 512 in CLIP, the additional traffic brought by transmitting $\phi_ n $ is extremely small compared with prompts. The server averages the parameters to obtain the aggregation model parameters:
\begin{linenomath}\begin{align}
\label{agg}
\mathbf{p}_g =\frac{1}{N} \sum_{n=1}^N \mathbf{p}_n,\quad \phi_g = \frac{1}{N} \sum_{n=1}^N \phi_n
\end{align}\end{linenomath}

Repeat Step 2 and Step 3 for $T$ rounds, then $\mathbf{p}_g$ and $\mathcal{Q}_g(\cdot)$ with parameters $\phi_g$ can be used to establish the global model. 

\begin{figure}[tb]
\centering
\includegraphics[width=0.88\linewidth]{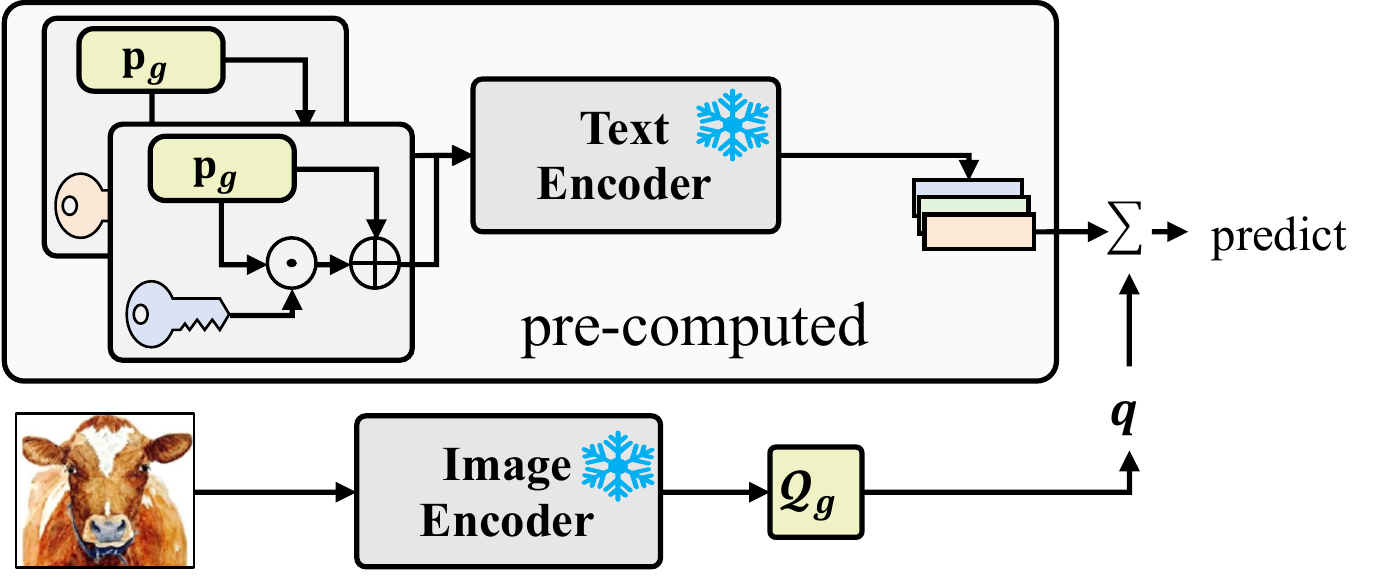}
\caption{Improve inference efficiency. }
\label{v2apt}
\end{figure}

\textbf{Improve inference efficiency.} Notably, during the inference process, assuming a batch size of $B$, since each sample is associated with distinct prompts, the text encoder needs to be executed $B$ times. This issue is also encountered in other visually conditioned prompt methods~\cite{zhou2022conditional}. To address this concern, we propose an enhanced strategy in Figure~\ref{v2apt}. By precomputing the text encodings for each key, we eliminate the need for redundant execution of the text encoder, leading to a significant reduction in inference costs.

\begin{figure}[tb]
\centering
\includegraphics[width=1.\linewidth]{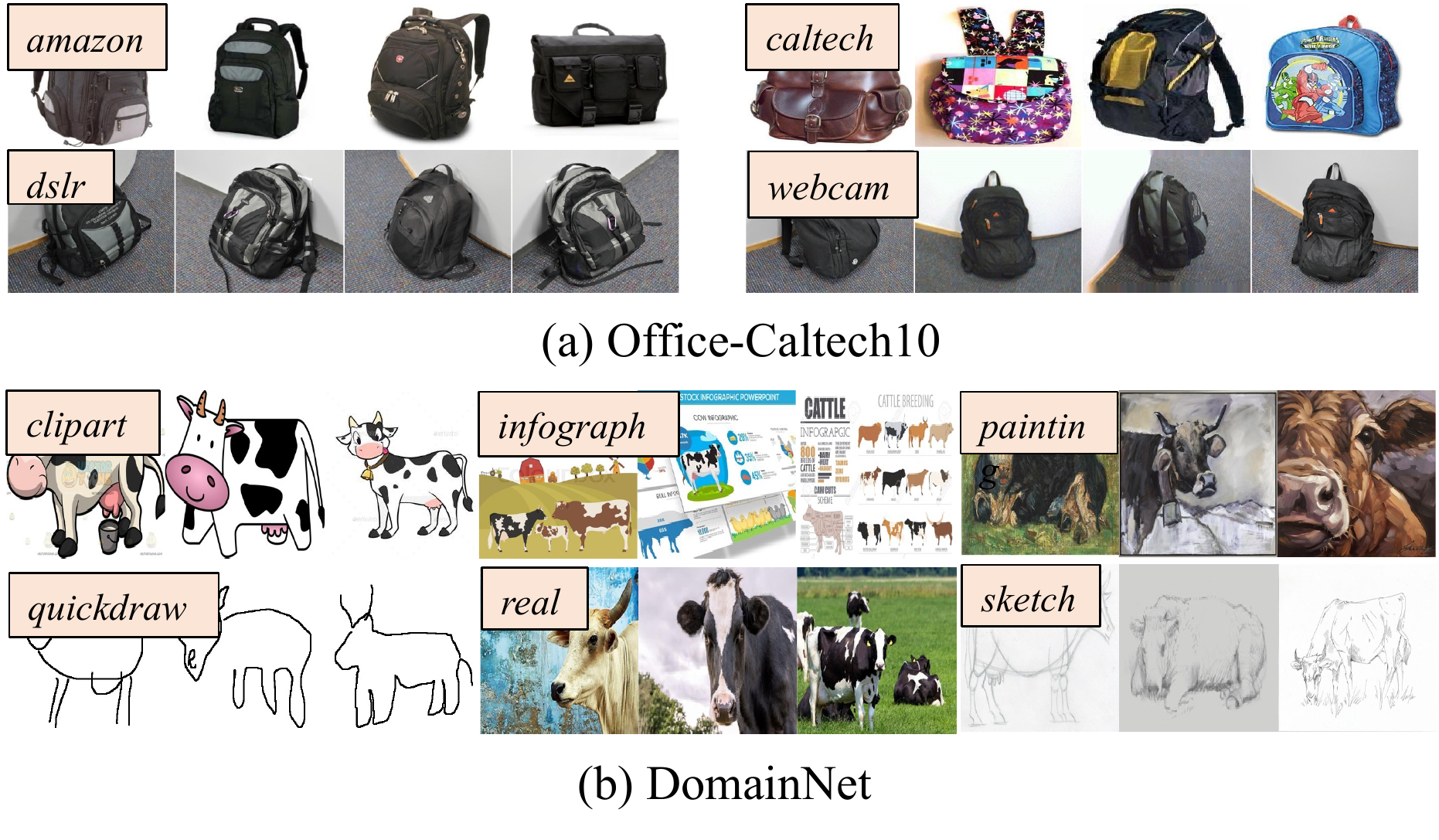}
\caption{(a) Office-Caltech10, which includes four domains. (b) DomainNet, which includes six domains.}
\label{dataset}
\end{figure}

\section{Experiments}
We conduct extensive experiments to evaluate the efficacy of FedAPT. We aim to answer the following questions: 1) Can we achieve better performance with fewer learnable parameters in FL using the pre-trained model CLIP, compared to a fully-trained ViT or ResNet? 2) How much improvement in global model can FedAPT achieve compared to existing federated tuning algorithms?

We first compare FedAPT with various baselines on two large-scale multi-domain image classification datasets. We consider two scenarios, where each domain acts as a client independently and each domain is divided into five non-IID clients. Then we conduct several ablation studies to observe the impact of keys on the client model and global model. We also explore the impact of different initialized key values on the performance of the global model.

\begin{table}[]
\center
\resizebox{1\linewidth}{!}{
\begin{tabular}{cl|c|cc|cccccc}
 \Xhline{1pt}  \rowcolor{gray}       
                                                                                                 &                    & zero-shot                                                & \multicolumn{2}{c|}{fully-trained}                                                                            & \multicolumn{6}{c}{fine-tune}                                                                                                                                                                            \\ \hline\cline{3-10} 
                                                              \rowcolor{gray}                             \multicolumn{2}{c|}{\begin{tabular}[c]{@{}c@{}}Datasets\\ \& Domains\end{tabular}  }  
   & \begin{tabular}[c]{@{}c@{}}CLIP\\ -zeroshot\end{tabular} & \begin{tabular}[c]{@{}c@{}}ResNet\\ -full\end{tabular} & \begin{tabular}[c]{@{}c@{}}ViT\\ -full\end{tabular} & \begin{tabular}[c]{@{}c@{}}ResNet\\ -tuning\end{tabular} & \begin{tabular}[c]{@{}c@{}}ViT\\ -tuning\end{tabular} & \begin{tabular}[c]{@{}c@{}}CLIP\\ -FC\end{tabular} &  \begin{tabular}[c]{@{}c@{}}Fed\\ -CLIP\end{tabular} &  \begin{tabular}[c]{@{}c@{}}Prompt\\ -FL\end{tabular}      & \begin{tabular}[c]{@{}c@{}}Fed\\ -APT\end{tabular}          \\\Xhline{0.8pt} 
\multicolumn{1}{c|}{\multirow{7}{*}{DN}}                                                  & \textit{c}   & 65.86                                                    & 32.44                                                  & 63.55                                               & 52.32                                                    & 71.93                                                 & 74.53                         & 70.25                       & 75.84          & \textbf{77.36}  \\
\multicolumn{1}{c|}{}                                                                            & \textit{i} & 40.50                                                    & 59.58                                                  & 27.07                                               & 20.85                                                    & 48.39                                                 & 48.00                         & 45.70                     & 49.82          & \textbf{52.17}  \\
\multicolumn{1}{c|}{}                                                                            & \textit{p}  & 62.25                                                    & 59.58                                                  & 49.00                                               & 44.66                                                    & 68.06                                                 & 69.08                           & 66.16                  & 70.81          & \textbf{72.51}  \\
\multicolumn{1}{c|}{}                                                                            & \textit{q} & 13.36                                                    & 43.25                                                  & 62.20                                               & 7.13                                                     & 21.24                                                 & 31.84                           & 16.98                   & 32.98          & \textbf{48.84}  \\
\multicolumn{1}{c|}{}                                                                            & \textit{r}      & 80.04                                                    & 74.73                                                  & 68.35                                               & 65.85                                                    & 80.43                                                 & 83.87                                              & 83.04  & 83.55          & \textbf{84.79}  \\
\multicolumn{1}{c|}{}                                                                            & \textit{s}    & 57.92                                                    & 61.81                                                  & 54.05                                               & 38.33                                                    & 63.45                                                 & 65.29                            & 61.75                   & 67.31          & \textbf{68.69}  \\ \cline{2-11} 
\multicolumn{1}{c|}{}                                                                            & avg                & 53.32                                                    & 55.23                                                  & 54.04                                               & 38.19                                                    & 58.92                                                 & 62.10                       &       57.31                & 63.39          & \textbf{67.39}  \\ \Xhline{0.8pt} 
\multicolumn{1}{c|}{\multirow{5}{*}{\begin{tabular}[c]{@{}c@{}}OC\end{tabular}}} & \textit{a}       & 95.69                                                    & 95.31                                                  & 36.97                                               & 95.31                                                    & \textbf{96.87}                                        & 96.35                               &     95.31          & 95.83          & 95.83           \\
\multicolumn{1}{c|}{}                                                                            & \textit{w}       & 93.24                                                    & 100.00                                                 & 36.20                                               & 98.27                                                    & 100.00                                                & 100.00                           &      98.27            & 98.27          & \textbf{100.00} \\
\multicolumn{1}{c|}{}                                                                            & \textit{d}       & 95.23                                                    & 100.00                                                 & 12.90                                               & 93.54                                                    & 96.77                                                 & 93.54                                              & 96.77  & 96.77          & \textbf{100.00} \\
\multicolumn{1}{c|}{}                                                                            & \textit{c}       & 92.54                                                    & 96.44                                                  & 29.77                                               & 93.77                                                    & 97.33                                                 & 96.00                                              & 96.00    & \textbf{98.22} & 96.44           \\ \cline{2-11} 
\multicolumn{1}{c|}{}                                                                            & avg             & 94.18                                                    & 97.94                                                  & 28.96                                               & 95.22                                                    & 97.74                                                 & 96.47                   &              96.58           & 97.27          & \textbf{98.07}  \\ \Xhline{0.8pt} 
\multicolumn{2}{c|}{\#params (M)}                                                                           & 0                                                       & 24.21                                                 & 87.86                                             & 0.71                                                   & 0.18                                                 & 0.26                 & 0.52                              & 2.826         & 2.829          \\ 
\Xhline{1pt}
\end{tabular}
}
\caption{ The accuracy of the global model in various domains in supervised setting with domain differences. The last row is the number of parameters to be transmitted in each round of communication. Overall, these results indicate that FedAPT achieves the best performance with few learnable parameters.}
\label{domainnet}
\end{table}

\begin{table}[t]
\center
\resizebox{1.0\linewidth}{!}{
\begin{tabular}{c|c|l|cccccc|c}
\Xhline{1pt} \rowcolor{gray}
                        &          $\beta$             &          & \textit{c} & \textit{i} & \textit{p} & \textit{q} & \textit{r}  & \textit{s} & \textbf{Average}         \\ \hline
\multirow{12}{*}{by domain} & \multirow{4}{*}{$0.01$}    & CLIP-FC  & 67.60  &44.39     & 65.32             & 22.97              & 79.42          & 59.66  & 56.56          \\
        			&                       & FedCLIP & \textbf{68.41}            &  \textbf{45.04  }            & 64.41            & 16.76              & \textbf{82.22 }          & \textbf{60.08}           & 56.15          \\
                        &                       & PromptFL & 66.68            & 43.75              & 65.31             & 24.45              & 77.75          & 58.74           & 56.11          \\
                        &                       & FedAPT   & 67.21            & 43.69              & \textbf{66.43}    & \textbf{26.70}     & 77.88 & 57.85           & \textbf{56.63 }\\ \cline{2-10} 
                        & \multirow{4}{*}{$0.5$}    & CLIP-FC  & 71.86            & 46.70              & 68.10             & 31.86              & 83.05          & 63.29           & 60.81          \\
                        &                       & FedCLIP & 69.54            & 45.61              & 65.15            & 18.00             & 82.61          & 60.39           & 56.88          \\
                        &                       & PromptFL & 73.19            & 47.72              & 69.59             & 37.31              & 82.98          & 64.85           & 62.61          \\
                        &                       & FedAPT   & \textbf{74.70}   & \textbf{49.94}     & \textbf{71.27}    & \textbf{46.89}     & \textbf{84.41} & \textbf{66.61}  & \textbf{65.64} \\ \cline{2-10} 
                        & \multirow{4}{*}{$5$}    & CLIP-FC  & 73.50            & 47.34              & 68.62             & 34.00              & 83.86          & 64.37           & 61.95          \\
                        &                       & FedCLIP &    69.37       &    45.61           &       65.58  &     18.67    &   82.77      &      60.73      &     57.12    \\
                        &                       & PromptFL & 74.49            & 48.34              & 69.77             & 38.30              & 83.94          & 66.05           & 63.48          \\
                        &                       & FedAPT   & \textbf{75.64}   & \textbf{50.63}     & \textbf{71.59}    & \textbf{50.55}     & \textbf{85.02} & \textbf{67.63}  & \textbf{66.84} \\ \Xhline{0.8pt} 
\multirow{12}{*}{by random}   & \multirow{4}{*}{$0.01$} & CLIP-FC  & 70.36  & 44.55     & \textbf{66.59}    & 21.10              & 80.03& \textbf{60.78}  & \textbf{57.24} \\
			&                               & FedCLIP & 68.04            & \textbf{45.10  }            & 63.45             &16.45              & \textbf{81.93 }          & 59.26           & 55.70          \\
                             &                               & PromptFL & 69.68            & 41.59              & 65.19             & 22.41              & 77.87          & 59.40           & 56.02          \\
                             &                               & FedAPT   & \textbf{70.50}            & 42.17              & 66.18             & \textbf{27.48}     & 77.22          & 59.41           & 57.16         \\ \cline{2-10} 
                             & \multirow{4}{*}{$0.5$}  & CLIP-FC  & 73.19            & 46.24              & 68.73             & 34.63              & 82.43          & 64.24           & 61.58          \\
                             &                               & FedCLIP & 69.86            & 45.26             & 65.50             & 17.46              & 82.65          & 61.05         &    56.96      \\
                              &                               & PromptFL & 73.71            & 46.13              & 69.53             & 39.65              & 82.10          & \textbf{65.22 }           & 62.72          \\
                             &                               & FedAPT   & \textbf{75.56}   & \textbf{48.41}     & \textbf{71.64}    & \textbf{47.28}     & \textbf{83.32} & 65.06 & \textbf{65.21} \\ \cline{2-10} 
                             & \multirow{4}{*}{$5$}    & CLIP-FC  & 74.19            & 47.00              & 67.85             & 31.77              & 84.08          & 65.88           & 61.80          \\
                             &                               & FedCLIP & 69.67  & 45.30             & 65.42    & 17.93              & 82.86          & 60.69         & 56.97         \\
                             &                               & PromptFL & 74.44  & 47.70              & \textbf{69.06}    & 38.68              & 84.66          & \textbf{67.51 }           & 63.68          \\
                             &                               & FedAPT   & \textbf{76.99 }            & \textbf{49.45}     & 68.49             & \textbf{49.53}     & \textbf{85.92} & 67.42  & \textbf{66.30 }
  \\ \Xhline{1pt}	 	 	 	  	 	 	 	  	 	 	 	 			 	 
\end{tabular}
}
\caption{The accuracy of the global model in various domains in supervised setting with domain and category differences. There are 30 clients in total. `by domain' means each domain randomly selects one client, and a total of six clients are selected in each round. `by random' means to randomly select six clients from all 30 clients. FedAPT's global model can significantly improve performance in various domains.}
\label{device}
\end{table}

\subsection{Experimental Setup}
\textbf{Datasets.}
We adopt two datasets, Office-Caltech10~\cite{gong2012geodesic} and DomainNet~\cite{peng2019moment}. \textbf{Office-Caltech10} (OC) is a small dataset that has four domains (\textit{amazon}, \textit{caltech}, \textit{dslr}, and \textit{webcam}), and each domain contains 10 overlapping categories between the Office-31~\cite{saenko2010adapting} dataset and the Caltech-256~\cite{griffin2007caltech} dataset. There are at least 100 and at most 800 images in different domains. Since the number of images is small, we use each domain as a client. \textbf{DomainNet} (DN) is a large-scale multi-domain dataset containing six domains (\textit{clipart}, \textit{infograph}, \textit{painting}, \textit{quickdraw}, \textit{real}, and \textit{sketch}), and each domain contains about 33k-120k images with 345 categories. We set the number of clients $N_{dom}$ in each domain as 1 or 5. When $N_{dom}=5$, we use the Dirichlet distribution to construct the non-IID property across clients. Both datasets contain 224$\times$224 pixel images. Figure~\ref{dataset} depicts the divergences of different domains.

\textbf{Compared Methods.}
To comprehensively evaluate the performance, we compare the following methods: 1) \textbf{ResNet-full}: Federated training with ResNet50~\cite{he2016deep}, all parameters participate in the training procedure. 2) \textbf{ResNet-tuning}, where the backbone of ResNet50 is frozen, only the last layer is trained. 3) \textbf{ViT-full} and 4) \textbf{ViT-tuning}: Replace ResNet in the above two baselines with ViT.
5) \textbf{CLIP-zeroshot} do not make any changes to the CLIP, and directly perform zero-shot image classification in all domains. This helps us understand how much the prompt tuning improves the CLIP.
6) \textbf{CLIP-FC}: Freeze the parameters of CLIP, and insert a learnable fully-connection layer at the end of the image encoder. Only the learnable layer is shared to the server.
7)  \textbf{FedCLIP}~\cite{lu2023fedclip}: Add an adapter at the end of the visual backbone and perform federated tuning. 8) \textbf{PromptFL}~\cite{guo2022promptfl}: A prompt tuning method for CLIP, details can be seen in Section~\ref{fltun}. 9) \textbf{FedAPT}: our proposed method. In the unsupervised experiment part, we utilize the local training method proposed in Section~\ref{trainsec} uniformly as FedCLIP and PromptFL have not yet explored federated tuning in an unsupervised setting.


\textbf{Implementation Details.} We use PyTorch to implement all methods. The SGD optimizer is used for both datasets.  We set Office-Caltech10 with a learning rate of 0.001 and batch size of 32, and DomainNet with a learning rate of 0.01 and batch size of 256. The global communication round $T_g$ is set to 50, and the local training epoch $T_l$ is set to $1$. The length of prompts $s$ is 16. In supervised setting, we set class-specific prompts $\mathbf{p}_g\in \mathbb{R}^{C\times s\times d}$.  In unsupervised setting, due to the lack of supervision information, we use the class-shared prompts $\mathbf{p}_g\in \mathbb{R}^{s\times d}$. The CLIP used in this paper takes ViT-B/32~\cite{dosovitskiy2020image} as the image encoder. Each experiment is repeated three times with different seeds, and the mean result is reported. All experiments are completed with one GeForce RTX 3090 GPU.

\subsection{Main Results}

\textbf{Supervised setting with domain differences. } We first treat each domain as a separate client with the same class distribution but different domain characteristics.

The results on the two datasets are reported in Table~\ref{domainnet}. We highlight the following points: \textbf{1)} From ResNet-full vs. ResNet-tuning and ViT-full vs. ViT-tuning, we can see that pre-trained ViT has more powerful representation ability than pre-trained ResNet. As a result, ResNet must be fully trained in order to achieve the accuracy rate obtained through ViT-tuning. \textbf{2)} CLIP's zero-shot classification capability is insufficient to meet the requirements of various domains. The performance gap with federated prompt tuning is large, which indicates that the original CLIP lacks knowledge of the client domains. \textbf{3)} Inserting a fully-connection layer after the image encoder of CLIP can remarkably enhance the performance on DomainNet, as indicated by CLIP-FC, which approaches the basic version of PromptFL. \textbf{4)} Considering the performance of different methods and the amount of communication parameters in each round, FedAPT achieves the highest performance at a very low cost, which shows its great potential in the application of pre-trained CLIP to FL scenarios.

\textbf{Supervised setting with domain and category differences.} We split each domain in DomainNet into five clients, i.e., sample $\boldsymbol{\pi}_c \sim \operatorname{Dir}\left(\beta \mathbf{1}_{5}\right)$ and allocate a $\boldsymbol{\pi}_{c,i}$ proportion of the instances with label $c$ to the training set of the $i$-th client in each domain. Smaller $\beta$ leads to larger data distribution differences among clients and more imbalanced classes. Finally, we obtain 30 clients. In addition to the divergences in data domains, there are also divergences in category distribution among clients. In each communication round, we randomly select one client from each domain, meaning six clients are selected for training. Note that because the random seeds are the same, the clients selected in each round by different methods are consistent.

The results are shown in Table~\ref{device}. As we can see, FedAPT still outperforms the competitors overall. Furthermore, it is worth noting that when dealing with extremely unbalanced categories, all CLIP tuning methods exhibit similar performance. This suggests that the pre-trained CLIP model needs to be augmented with additional designs to achieve better performance, as its existing capabilities may be insufficient.

 \begin{table}[]
 \center
\resizebox{1\linewidth}{!}{	
\begin{tabular}{ccccccccc}
\Xhline{1pt}\rowcolor{gray}
                                                                                              & \multicolumn{1}{c|}{}         & \textit{c }    & \textit{i }    & \textit{p}     & \textit{q}     & \textit{r}     & \multicolumn{1}{c|}{\textit{s}}     & Average \\ \hline
\multicolumn{9}{c}{domain differences}                                                                                                                                                                       \\ \hline
\multicolumn{2}{c|}{FedCLIP}                                                                                                  & 68.23 & 45.14 & 64.71 & 18.07 & 82.00 & \multicolumn{1}{c|}{60.00} & 56.36   \\
\multicolumn{2}{c|}{PromptFL}                                                                                                 & 68.37 & 46.61 & 65.67 & 17.34 & 82.17 & \multicolumn{1}{c|}{60.49} & 56.78   \\
\multicolumn{2}{c|}{FedAPT}                                                                                                   & 69.48 & 47.18 & 66.12 & 17.89 & 83.09 & \multicolumn{1}{c|}{61.19} & \textbf{57.49}   \\ \hline
\multicolumn{9}{c}{domain and category differences}                                                                                                                                                          \\ \hline
\multicolumn{1}{c|}{\multirow{3}{*}{\begin{tabular}[c]{@{}c@{}}$\beta$\\ =0.01\end{tabular}}} & \multicolumn{1}{c|}{FedCLIP}  & 67.65 & 43.33 & 63.48 & 17.55 & 81.33 & \multicolumn{1}{c|}{59.17} & 55.42   \\
\multicolumn{1}{c|}{}                                                                         & \multicolumn{1}{c|}{PromptFL} & 68.31 & 44.41 & 64.31 & 16.37 & 81.29 & \multicolumn{1}{c|}{60.21} & 55.82   \\
\multicolumn{1}{c|}{}                                                                         & \multicolumn{1}{c|}{FedAPT}   &  68.23 & 44.99 & 64.29 & 16.70 & 81.57 & \multicolumn{1}{c|}{60.17} & \textbf{55.99}          \\ \hline
\multicolumn{1}{c|}{\multirow{3}{*}{\begin{tabular}[c]{@{}c@{}}$\beta$\\ =0.5\end{tabular}}}  & \multicolumn{1}{c|}{FedCLIP}  & 67.59 & 44.11 & 63.68 & 18.21 & 81.52 & \multicolumn{1}{c|}{59.06} & 55.70   \\
\multicolumn{1}{c|}{}                                                                         & \multicolumn{1}{c|}{PromptFL} & 68.75 & 46.22 & 65.13 & 17.68 & 82.90 & \multicolumn{1}{c|}{60.62} & 56.88   \\
\multicolumn{1}{c|}{}                                                                         & \multicolumn{1}{c|}{FedAPT}   & 69.02 & 46.67 & 65.88 & 18.31  & 83.00 & \multicolumn{1}{c|}{60.96} & \textbf{57.31}         \\ \hline
\multicolumn{1}{c|}{\multirow{3}{*}{\begin{tabular}[c]{@{}c@{}}$\beta$\\ =5\end{tabular}}}    & \multicolumn{1}{c|}{FedCLIP}  & 67.81 & 44.32 & 63.82 & 18.16 & 81.58 & \multicolumn{1}{c|}{59.20} & 55.82   \\
\multicolumn{1}{c|}{}                                                                         & \multicolumn{1}{c|}{PromptFL} &  68.64       &  46.27      &  65.60      &   17.52     &   82.50    & \multicolumn{1}{c|}{60.38}      &   56.82       \\
\multicolumn{1}{c|}{}                                                                         & \multicolumn{1}{c|}{FedAPT}   & 69.13       &  47.34     &  66.41      &   18.47     &   83.31    & \multicolumn{1}{c|}{61.58}      &   \textbf{57.71}       \\ \Xhline{1pt}
\end{tabular}}
\caption{The accuracy of the global model in various domains in unsupervised settings.} 	 	  	 	 					 	 	
				 	 		 	 	 	 	 	 	
\label{unsuper}
\end{table}

\textbf{Unsupervised setting.} We replicate the above experiments in an unlabeled setting, and the results are shown in Table~\ref{unsuper}. Specifically, we observe that \textbf{1)} On DomainNet dataset, the performance of unsupervised federated tuning is comparable to that of supervised tuning. This indicates that the unsupervised fine-tuning method proposed in Section~\ref{fltun} is effective. \textbf{2)} Under the unsupervised setting, FedAPT still outperforms the baseline methods, demonstrating its significant effectiveness. In addition, as we use the class-shared prompts for FedAPT and PromptFL in this setting, the communication parameter sizes per round are 0.011M and 0.008M, respectively. This demonstrates the advantage of federated prompt tuning in terms of communication costs.

\subsection{Ablation Studies}
\begin{table}[]
\center
\resizebox{1\linewidth}{!}{	
\begin{tabular}{lccccccc}
\Xhline{1pt}  \rowcolor{gray}
      & \textit{c} & \textit{i} & \textit{p} & \textit{q} & \textit{r}  & \textit{s} & Average\\ \Xhline{1pt} 
meta prompt $\mathbf{p}_g$    & 74.62            & 48.04              & 68.82             & 22.18              & 82.71          & 65.98        &60.39   \\ \hline
$\mathbf{p}_g+\mathbf{p}_g\odot \mathbf{e}^{\prime}_0$   & \textbf{78.76}   & 44.77              & 65.85             & 20.47              & 80.80          & 63.85           &59.08 \\
$\mathbf{p}_g+\mathbf{p}_g\odot \mathbf{e}^{\prime}_1$  & 70.96            & \textbf{52.70}     & 65.83             & 19.26              & 81.03          & 62.40         &58.70 \\
$\mathbf{p}_g+\mathbf{p}_g\odot \mathbf{e}^{\prime}_2$  & 71.98            & 46.05              & \textbf{74.58}    & 16.63              & 81.12          & 61.20      &58.59      \\
$\mathbf{p}_g+\mathbf{p}_g\odot \mathbf{e}^{\prime}_3$   & 69.66            & 39.99              & 60.27             & \textbf{48.84}     & 75.78          & 58.67        &58.87    \\
$\mathbf{p}_g+\mathbf{p}_g\odot \mathbf{e}^{\prime}_4$  & 73.70            & 47.56              & 68.12             & 19.66              & \textbf{85.94} & 64.41      &59.90      \\
$\mathbf{p}_g+\mathbf{p}_g\odot \mathbf{e}^{\prime}_5$  & 72.28            & 44.92              & 65.16             & 20.63              & 81.04          & \textbf{70.42} &59.08  \\ \hline
Eq.~\ref{eq.pg} &\textit{ \textbf{77.36}  }         & \textit{ \textbf{52.17 }  }               & \textit{ \textbf{72.51}  }               & \textit{ \textbf{48.84 }  }               & \textit{ \textbf{84.79 }  }           & \textit{ \textbf{68.69  }  }       &\textit{ \textbf{67.39}  }       \\  \Xhline{1pt} 
\end{tabular}
}
\caption{The impact of combining meta prompts with keys of different domains.}
\label{key}
\end{table}


\begin{table}[t]
\center
\resizebox{.8\columnwidth}{!}{
\begin{tabular}{l|cccccc}
\Xhline{1pt} \rowcolor{gray}
                     & \textit{c} & \textit{i} & \textit{p} & \textit{q} & \textit{r}  & \textit{s}  \\ \hline
w/o key & 78.30            & 51.48              & 73.99             & 49.05              & 85.35         & 69.72           \\
w key   & 78.03            & 51.59              & 74.00             & 49.35              & 85.67         & 69.73        \\ \Xhline{1pt}
\end{tabular}}
\caption{The impact of keys in local training. We show that adding the keys does not destroy the local training. }
\label{localmodel}
\end{table}


\begin{table}[t]
\resizebox{0.98\columnwidth}{!}{
\begin{tabular}{cc|ccccc|c}
\Xhline{1pt}\rowcolor{gray}
                                                  &           & \textit{i}     & \textit{p}      & \textit{q}      & \textit{r}      & \textit{s}      & unseen domain (\textit{c} ) \\ \hline
\multicolumn{2}{c|}{PromptFL}                                  & 49.71 & 71.18 & 34.09 & 83.58 & 67.26 & 70.60         \\ \hline
\multicolumn{1}{c|}{\multirow{2}{*}{FedAPT }} & $\tau=0.01$ & 52.23 & 72.76 & 51.25 & 84.67 & 69.35 & 68.26          \\
\multicolumn{1}{c|}{}                             & $\tau=1$    & 52.27 & 72.14 & 51.05 & 84.18 & 68.77 & 70.50          \\ \Xhline{1pt}
\end{tabular}}
\caption{The impact of the adaptive network on seen and unseen domains.}
\label{unseen}
\end{table}

\textbf{Impact of keys.} We do a comprehensive study of the role of `key'. We disassemble the global model trained with FedAPT in Table~\ref{domainnet} and compare its performance under different situations: 1) Use the learned meta prompt $\mathbf{p}_g$ to test without any key. 2) Use keys from different domains for testing, such as $\mathbf{p}_g+\mathbf{p}_g\odot \mathbf{e}^{\prime}_k$.
3) Use the complete APT module, i.e., Eq.~\ref{eq.pg}. The results are reported in Tabe~\ref{key}. The classification ability of the meta prompt $\mathbf{p}_g$ is shown in the first row. Because there is no suitable key, its performance is mediocre. Then the middle six rows show the performance of $\mathbf{p}_g$ with different keys. We can see that the $k$-th key $\mathbf{e}^{\prime}_k$ can significantly improve the accuracy in the $k$-th domain.

 \textbf{The affect of the adaptive network.} 
 We evaluate the performance of the global model in both client domains and unseen new domains. We introduce softmax temperature $\tau$ to the output of the adaptive network, and the performance at $\tau=0.01$ can be interpreted as directly determining ``which key should be used". The results in Table~\ref{unseen} show that directly selecting a single key ($\tau=0.01$) improves the performance in client domains but sacrifices generalization in unseen domains. By adjusting $\tau$, we can significantly enhance client performance while maintaining nearly unchanged generalization performance in unknown domains.
 				


\textbf{Other results.} \textbf{1)} One worrying point about using the key on the client side is whether the key will limit the learning ability of the local model. We verify this point with an experiment. We repeat the experiment in Table~\ref{domainnet} and report the performance of local models from different domains in Table~\ref{localmodel}. The results show that the use of the key does not limit the learning of the local model. \textbf{2)} We also compare different initialized keys.  The results suggest that the keys work well for different types of random initialization, including  the uniform and the standard normal distribution initialization and random orthogonal initialization. We provide the numerical results in the supplementary materials. \textbf{3)} After precomputing and saving the text branch outputs of different keys, only the image branch and domain classification weights need to be computed during inference.  The final inference cost is as follows: CLIP's GFLOPs is 4.42 while that of ResNet50 is 4.14.  \textbf{4)} We demonstrate the impact in inference efficiency of the strategy in Figure~\ref{v2apt}. While maintaining nearly unchanged accuracy, the inference time has been reduced from thirty minutes to 30 seconds in one GPU card. Detailed numerical results are provided in the supplementary materials.

\section{Conclusions}

In this paper, we apply the pre-trained CLIP to the multi-domain federated learning in both supervised and unsupervised settings, and propose an adaptive prompt tuning method that uses domain-specific keys to generate specific prompts for each test sample. We have extensively validated the effectiveness of FedAPT. With less than 10\% of learnable parameters, FedAPT achieves performance surpassing that of fully-trained models. Moreover, when faced with challenges such as feature and category differences in client data, FedAPT demonstrates a significant performance improvement compared with the competitors.

\section{Acknowledgements}
This work was supported in part by the National Key R\&D Program of China (No.2021ZD0112803), the National Natural Science Foundation of China (No.62176061), STCSM project (No.22511105000), the Shanghai Research and Innovation Functional Program (No.17DZ2260900), and the Program for Professor of Special Appointment (Eastern Scholar) at Shanghai Institutions of Higher Learning.

\bibliography{aaai24}

\clearpage
\appendix

\section{Details of Experiments}
\begin{figure}[th!]
\centering
\includegraphics[width=0.9\linewidth]{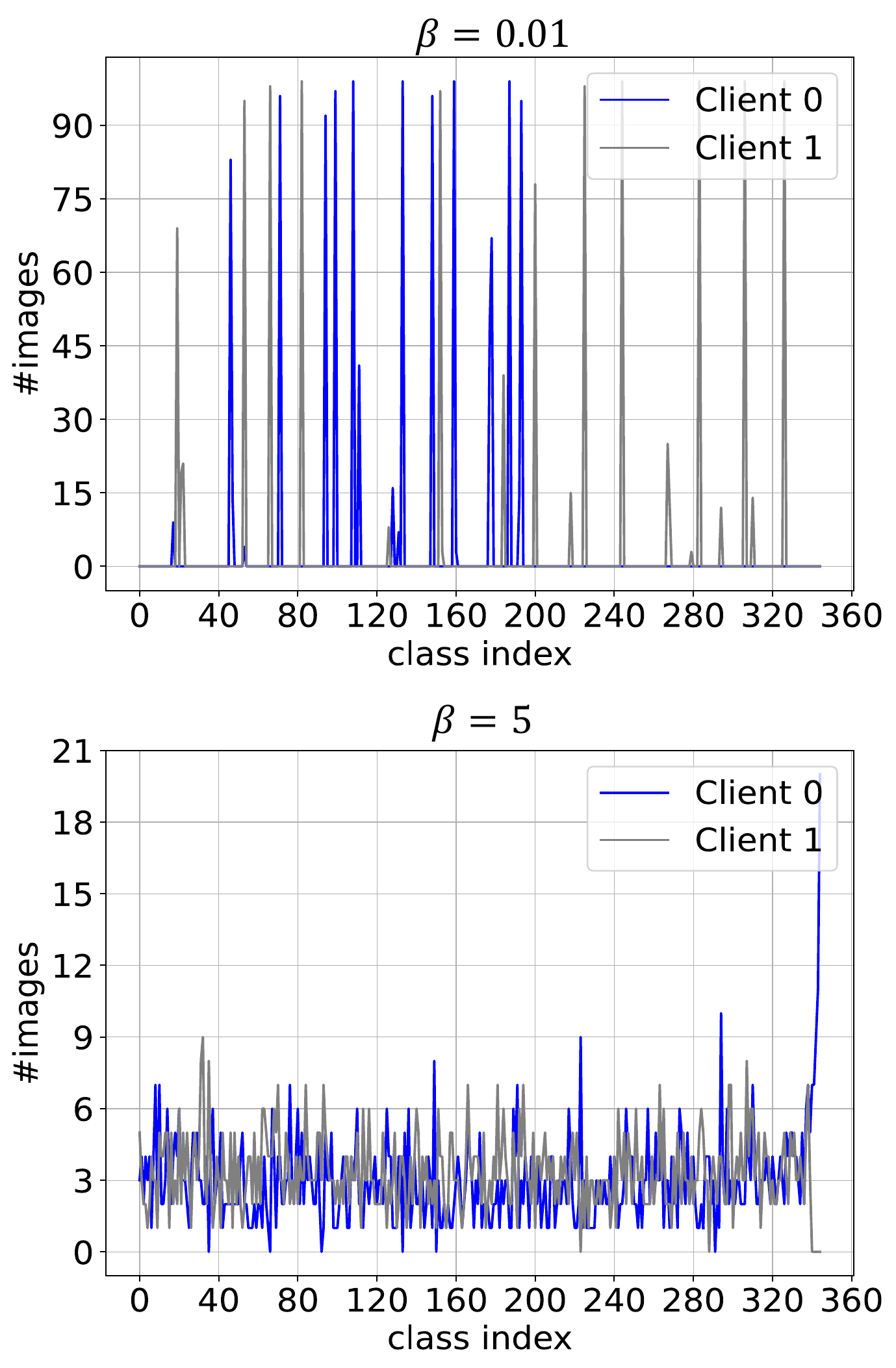}
\caption{Unbalanced label distribution controlled by $\beta$. When $\beta$ is small, there is almost no category overlap between different clients.}
\label{par}
\end{figure}

In some of the experiments, we utilize a standard Dirichlet distribution to partition the dataset, thereby creating disparities in class distributions among clients. Figure~\ref{par} illustrates the impact of different hyperparameter values, denoted as $\beta$, on the divergence in class distributions. As $\beta$ decreases, the disparities increase, and when $\beta$ approaches 0, there is minimal class overlap among clients.

We use the standard pre-trained CLIP. All images is reshaped into 224*224. We use the preprocess pipeline defined by CLIP, which includes Resize, CenterCrop, ToTensor and Normalize.

The details of the local datasets when each domain has one client are shown in Table~\ref{data1}. When each domain has five clients, the training set of each domain is divided into 5 parts.

\begin{table}[th!]
\center
\resizebox{1.0\columnwidth}{!}{
\begin{tabular}{c|cccccc}
\Xhline{1pt}  \rowcolor{gray}
      & \begin{tabular}[c]{@{}c@{}}Client0 \\ clipart\end{tabular} & \begin{tabular}[c]{@{}c@{}}Client1 \\ infograph\end{tabular} & \begin{tabular}[c]{@{}c@{}}Client2\\  painting\end{tabular} & \begin{tabular}[c]{@{}c@{}}Client3 \\ quickdraw\end{tabular} & \begin{tabular}[c]{@{}c@{}}Client3 \\ real\end{tabular} & \begin{tabular}[c]{@{}c@{}}Client5 \\ sketch\end{tabular} \\ \hline
train & 33k                                                        & 36k                                                          & 50k                                                         & 120k                                                         & 121k                                                    & 48k                                                       \\
test  & 15k                                                        & 16k                                                          & 22k                                                         & 52k                                                          & 52k                                                     & 21k                                                       \\ \Xhline{1pt}
\end{tabular}}
\caption{The local datasets when each domain has one client.}
\label{data1}
\end{table}

\section{Additional Experiments}
\subsubsection{Initialization of keys} We also compare different initialized keys. Table~\ref{randinit} shows the results. `rand\_U' means the uniform distribution U(0,1). `rand\_N' means the standard normal distribution.  `rand\_01' means to randomly generate one 01 matrix. `rand\_O' indicates random orthogonal initialization, that is, the keys of each domain are orthogonal. We can see that the results of different initializations are stable.

\subsubsection{The improve of inference efﬁciency} We demonstrate the impact in inference efﬁciency of the strategy in Figure 3 of the main paper. 
By precomputing text encodings for different keys, we avoid the need to compute text encodings for each sample in a batch repeatedly, thus saving a considerable amount of inference cost.
While maintaining nearly unchanged accuracy, the inference time has been reduced by about 60 times, see Table~\ref{times}.

\begin{table}[th!]
\center
\begin{tabular}{l|llllll}
\Xhline{1pt} \rowcolor{gray}
               &  \textit{c} & \textit{i} & \textit{p} & \textit{q} & \textit{r}  & \textit{s}  \\ \hline
original (min) & 19   & 39 & 31 & 30 & 60 & 30 \\
Improved (sec) & 19.5 & 26 & 30 & 65 & 67 & 32 \\ \Xhline{1pt} 
\end{tabular}
\caption{The improve of inference efﬁciency.}
\label{times}
\end{table}

\begin{table}[th!]
\center
\resizebox{1.\columnwidth}{!}{
\begin{tabular}{lccccccc}
\Xhline{1pt} \rowcolor{gray}
         & \textit{c} & \textit{i} & \textit{p} & \textit{q} & \textit{r}  & \textit{s} & Average \\ \hline
rand\_U  & 77.36   & 52.17     & 72.51    & 48.84     & 84.79 & 68.69  & 67.39   \\
rand\_N  & 75.78   & 51.86     & 72.02    & 53.82     & 84.20 & 67.40  & 67.51   \\
rand\_01 & 76.60   & 52.33     & 72.59    & 52.89     & 84.83 & 68.30  & 67.92   \\
rand\_O  & 77.00   & 52.25     & 72.83    & 51.29     & 84.86 & 68.64  & 67.81   \\  \Xhline{1pt}
\end{tabular}}
\caption{The effect of different initialization keys. It shows the robustness of initialization methods to keys.}
\label{randinit}
\end{table}

\begin{table*}[t!]
\center
\resizebox{.8\linewidth}{!}{
\begin{tabular}{clcccccccc}
\Xhline{1pt}  \rowcolor{gray}
\multicolumn{1}{c|}{zero-shot}                                                                      & \multicolumn{1}{l|}{CLIP-zeroshot} & 65.86            & 40.50              & 62.25             & 13.36              & 80.04         & \multicolumn{1}{c|}{57.92}           & \multicolumn{1}{c|}{53.32}          & 0M                                                                \\ \hline
\multicolumn{1}{c|}{\multirow{2}{*}{fully-trained}}                                                 & \multicolumn{1}{l|}{ResNet-full}   & 71.74            & 32.44              & 59.58             & 43.25              & 74.73         & \multicolumn{1}{c|}{61.81}           & \multicolumn{1}{c|}{57.26}          & 24.21M                                                            \\
\multicolumn{1}{c|}{}                                                                               & \multicolumn{1}{l|}{ViT-full}      & 63.55            & 27.07              & 49.00             & 62.20              & 68.35         & \multicolumn{1}{c|}{54.05}           & \multicolumn{1}{c|}{54.04}          & 87.86M                                                            \\ \hline
\multicolumn{1}{c|}{\multirow{3}{*}{\begin{tabular}[c]{@{}c@{}}fine-tune \\ fc layer\end{tabular}}} & \multicolumn{1}{l|}{ResNet-tuning} & 52.32            & 20.85              & 44.66             & 7.13               & 65.85         & \multicolumn{1}{c|}{38.33}           & \multicolumn{1}{c|}{38.19}          & 0.71M                                                             \\
\multicolumn{1}{c|}{}                                                                               & \multicolumn{1}{l|}{ViT-tuning}    & 71.93            & 48.39              & 68.06             & 21.24              & 80.43         & \multicolumn{1}{c|}{63.45}           & \multicolumn{1}{c|}{58.92}          & 0.18M                                                             \\
\multicolumn{1}{c|}{}                                                                               & \multicolumn{1}{l|}{CLIP-fc}       & 74.53            & 48.00              & 69.08             & 31.84              & 83.87         & \multicolumn{1}{c|}{65.29}           & \multicolumn{1}{c|}{62.10}          & 0.26 M                                                            \\ \hline
\multicolumn{1}{c|}{\multirow{9}{*}{\begin{tabular}[c]{@{}c@{}}prompt \\ tuning\end{tabular}}}      & \multicolumn{9}{c}{$s_p=8$}                                                                                                                                                                                                                                                          \\ \cline{2-10} 
\multicolumn{1}{c|}{}                                                                               & \multicolumn{1}{l|}{PromptFL}      & 75.95            & 49.60              & 70.73             & 31.06              & 83.77         & \multicolumn{1}{c|}{67.12}           & \multicolumn{1}{c|}{63.04}          & 1.413M                                                            \\
\multicolumn{1}{c|}{}                                                                               & \multicolumn{1}{l|}{FedAPT}        & 77.28            & 52.07              & 72.45             & 47.31              & 84.76         & \multicolumn{1}{c|}{68.73}           & \multicolumn{1}{c|}{67.10}          & 1.416M                                                            \\ \cline{2-10} 
\multicolumn{1}{c|}{}                                                                               & \multicolumn{9}{c}{$s_p=16$}                                                                                                                                                                                                                                                         \\ \cline{2-10} 
\multicolumn{1}{c|}{}                                                                               & \multicolumn{1}{l|}{PromptFL}      & 75.84            & 49.82              & 70.81             & 32.98              & 83.55         & \multicolumn{1}{c|}{67.31}           & \multicolumn{1}{c|}{63.39}          & 2.836M                                                            \\
\multicolumn{1}{c|}{}                                                                               & \multicolumn{1}{l|}{FedAPT}        & 77.36            & 52.17              & 72.51             & 48.84              & 84.79         & \multicolumn{1}{c|}{68.69}           & \multicolumn{1}{c|}{67.39}          & 2.829M                                                            \\ \cline{2-10} 
\multicolumn{1}{c|}{}                                                                               & \multicolumn{9}{c}{$s_p=24$}                                                                                                                                                                                                                                                         \\ \cline{2-10} 
\multicolumn{1}{c|}{}                                                                               & \multicolumn{1}{l|}{PromptFL}      & 76.11            & 49.77              & 70.62             & 32.47              & 83.46         & \multicolumn{1}{c|}{67.38}           & \multicolumn{1}{c|}{63.30}          & 4.239M                                                            \\
\multicolumn{1}{c|}{}                                                                               & \multicolumn{1}{l|}{FedAPT}        & 77.34            & 52.29              & 72.87             & 50.35              & 84.80         & \multicolumn{1}{c|}{68.54}           & \multicolumn{1}{c|}{\textbf{67.70}} & 4.242M                                                            \\ \Xhline{1pt}
\end{tabular}}
\caption{The performance of the global model. $s_p$ denotes the length of prompts. }
\label{sup1}
\end{table*}

\begin{figure}[th!]
\centering
\includegraphics[width=0.6\linewidth]{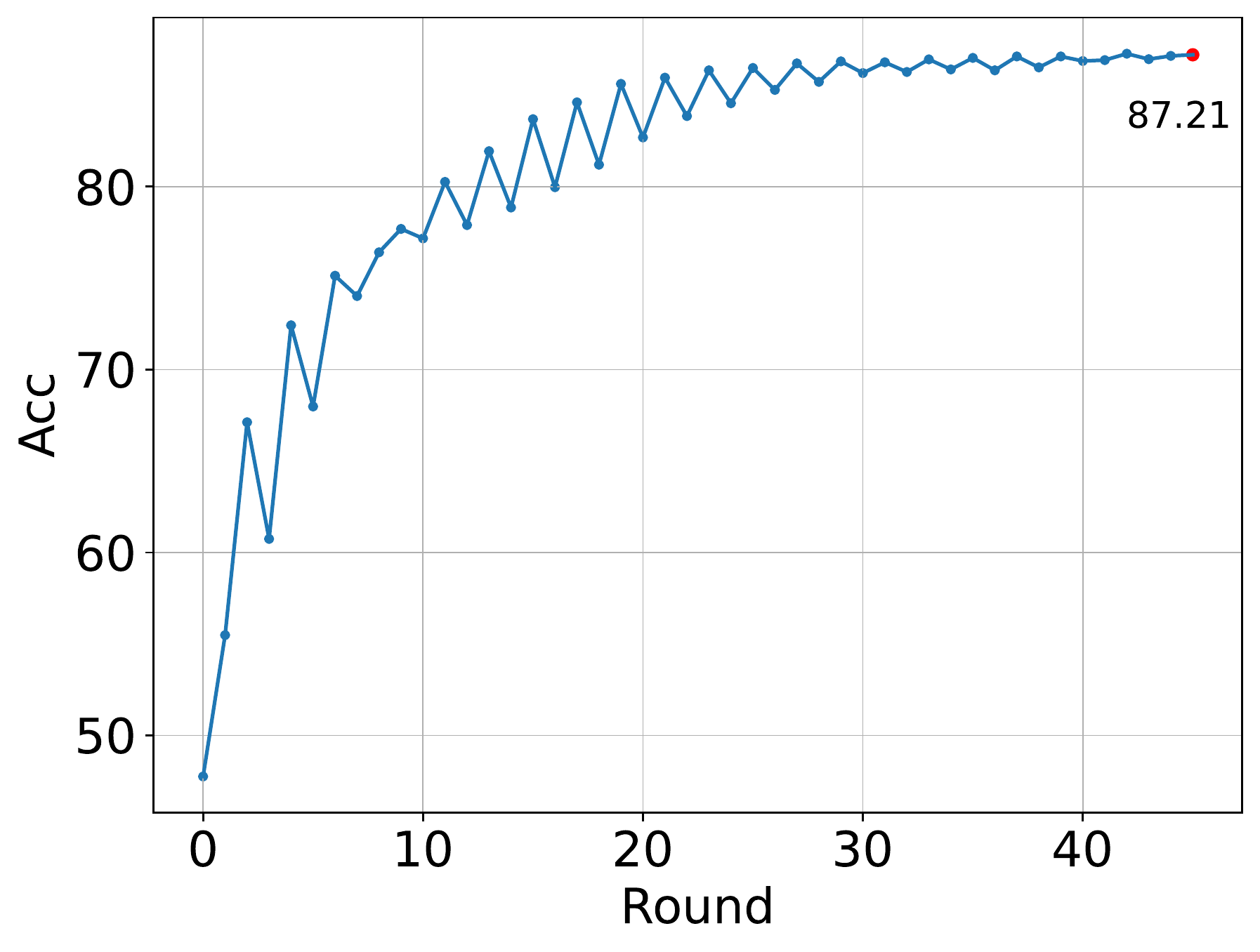}
\caption{The adaptive network's accuracy in classifying domains.}
\label{qacc}
\end{figure}

\subsubsection{Influence of the length of prompts.}
In the main paper, we set the length of prompts to $16$ for PromptFL and FedAPT.  Here we show the performance of the global model obtained by using prompts of different lengths. From Table~\ref{sup1}, we can see that when the length of prompts increases, the performance of FedAPT will also increase, and FedAPT still outperforms all comparison methods. Furthermore, the performance of PromptFL is not improved with the increase of $s_p$.

\subsubsection{Performance of $\mathcal{Q}$.}
We take the test set of all domains as a whole, label the data according to the domain index, and test the classification accuracy of the adaptive network. The result is as shown in Figure~\ref{qacc}. The adaptive network can converge rapidly in the early communication period.

\end{document}